\documentclass[10pt,twocolumn,letterpaper]{article}
\usepackage{iccv}
\usepackage{times}
\usepackage{epsfig}
\usepackage{graphicx}
\usepackage{amsmath}
\usepackage{amssymb}
\usepackage{pifont}
\usepackage{tabu}

\usepackage[pagebackref=true,breaklinks=true,letterpaper=true,colorlinks,bookmarks=false]{hyperref}

\usepackage{url}            
\usepackage{booktabs}       
\usepackage{amsfonts}       
\usepackage{nicefrac}       
\usepackage{microtype}      

\usepackage{tabu}           
\usepackage{multirow}
\usepackage{subcaption}
\usepackage{bigstrut}
\usepackage{amsmath}
\usepackage{soul}
\usepackage{algorithm}
\usepackage{algorithmicx}
\usepackage{algpseudocode}
\usepackage{adjustbox}
\usepackage{bm}

\DeclareMathOperator*{\argmin}{arg\,min}

\def \ours {DIML}
\newcommand{\cmark}{\ding{51}}%
\newcommand{\paragrapha}[2]{\vspace{#1}\noindent\textbf{#2}}

\DeclareMathOperator{\GAP}{GAP}
\DeclareMathOperator*{\arginf}{\arg\inf}
\newcommand{\sss}{\mathrm{s}}
\newcommand{\ttt}{\mathrm{t}}
\newcommand{\structural}{{\rm struct}}

\iccvfinalcopy 

\def\iccvPaperID{2020} 

\ificcvfinal\fi

\begin{document}

\title{Towards Interpretable Deep Metric Learning with Structural Matching}

\author{Wenliang Zhao\textsuperscript{1,2,3}\thanks{Equal contribution.}, ~Yongming Rao\textsuperscript{1,2,3}\footnotemark[1], ~Ziyi Wang\textsuperscript{1,2,3}, ~Jiwen Lu\textsuperscript{1,2,3}\thanks{Corresponding author.}, ~Jie Zhou\textsuperscript{1,2,3}\\
\textsuperscript{1}Department of Automation, Tsinghua University, China\\
\textsuperscript{2}State Key Lab of Intelligent Technologies and Systems, China\\
\textsuperscript{3}Beijing National Research Center for Information Science and Technology, China\\
{\tt\small zhaowl20@mails.tsinghua.edu.cn; raoyongming95@gmail.com; }\\ 
{\tt\small wziyi20@mails.tsinghua.edu.cn;  \{lujiwen, jzhou\}@tsinghua.edu.cn} \\
}

\maketitle
\ificcvfinal\thispagestyle{empty}\fi

\begin{abstract}
  How do the neural networks distinguish two images? It is of critical importance to understand the matching mechanism of deep models for developing reliable intelligent systems for many risky visual applications such as surveillance and access control. However, most existing deep metric learning methods match the images by comparing feature vectors, which ignores the spatial structure of images and thus lacks interpretability. In this paper, we present a deep interpretable metric learning (DIML) method for more transparent embedding learning. Unlike conventional metric learning methods based on feature vector comparison, we propose a structural matching strategy that explicitly aligns the spatial embeddings by computing an optimal matching flow between feature maps of the two images. Our method enables deep models to learn metrics in a more human-friendly way, where the similarity of two images can be decomposed to several part-wise similarities and their contributions to the overall similarity. Our method is model-agnostic, which can be applied to off-the-shelf backbone networks and metric learning methods. We evaluate our method on three major benchmarks of deep metric learning including CUB200-2011, Cars196, and Stanford Online Products, and achieve substantial improvements over popular metric learning methods with better interpretability. Code is available at \url{https://github.com/wl-zhao/DIML}.
\end{abstract}


\section{Introduction}

\begin{figure}[th]
    \centering
    \includegraphics[width=0.48\textwidth]{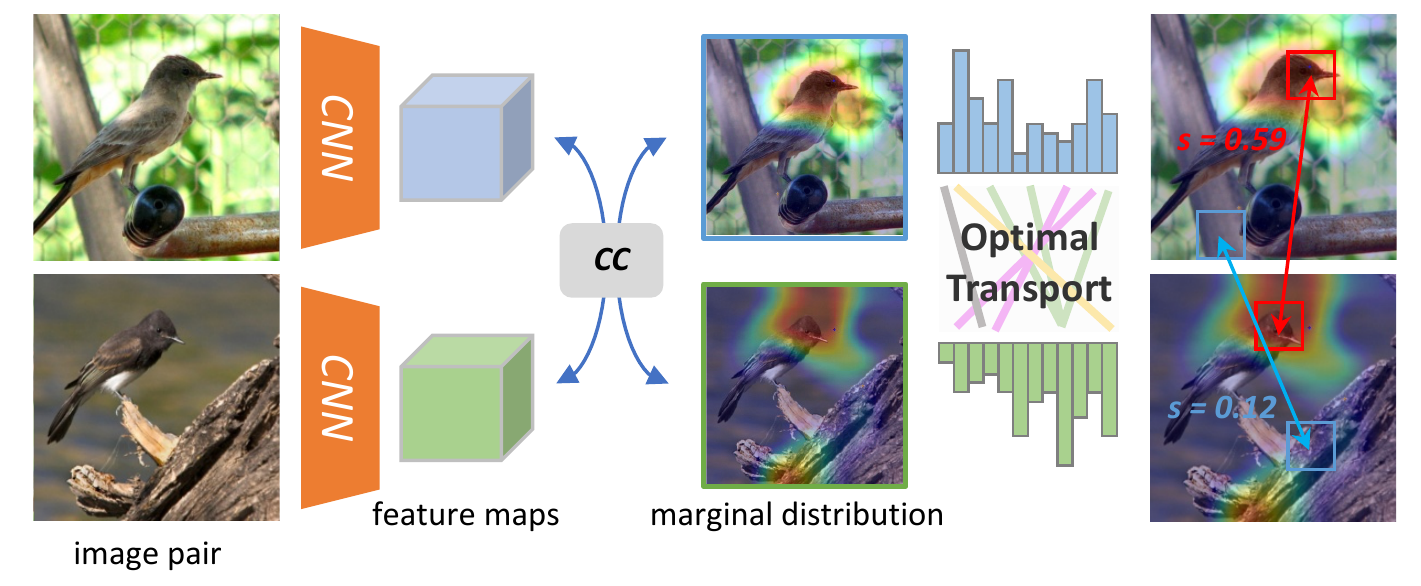}
    \vspace{-4pt}
    \caption{The main idea of the proposed deep interpretable metric learning (DIML) method. Unlike most existing deep metric learning methods match the images by comparing feature vectors, we propose a structural matching strategy that explicitly aligns the spatial embeddings by computing an optimal matching flow between feature maps of the two images to improve the interpretability of visual similarity.}
    \label{fig:intro}
    \vspace{-10pt}
\end{figure}

Visual similarity plays an important role in a range of vision tasks including image retrieval~\cite{sohn2016improved_npair}, person identification~\cite{Beyond-triplet-loss-reid} and image clustering~\cite{schroff2015facenet}. Recent advances in learning visual similarity are mostly driven by Deep Metric Learning (DML), which leverages deep neural networks to learn an embedding space where the embedding similarity in this space can meaningfully reflect the semantic similarity between samples. A variety of deep metric learning methods have been proposed and have shown strong superiority in learning accurate and generalizable visual similarities on various tasks~\cite{deng2019arcface,wang2019multi,kim2020proxy}. Despite the great progress in learning discriminative embeddings, deep metric learning methods with better interpretability have drawn limited attention from the community.  Understanding the underlying matching mechanism of deep metric learning models is of critical importance for developing reliable intelligent systems for many risky visual applications such as surveillance~\cite{sreenu2019intelligent} and access control~\cite{masi2018deep}. 

To improve the transparency of deep visual models, many efforts have been made recently by either explaining the existing models~\cite{cam,gradcam,Gan-dissection,transformerInterpretability} or modifying models to achieve better interpretability~\cite{zhang2018interpretable,zhang2019interpretingtree}. For example, visual attribution methods leverage correlation or gradient to find the important regions that have high contributions to the final prediction. \cite{zhang2018interpretable} and \cite{zhang2019interpretingtree} propose to add part constraints and tree structures to construct interpretable CNN models respectively. However, these methods are only designed for explaining the reasoning process of how the output of a deep model is produced and did not consider the interaction between samples. Although they achieve promising results on image classification~\cite{cam,transformerInterpretability}, visual question answering~\cite{gradcam}, and image generation~\cite{Gan-dissection}, they cannot explain how visual similarity is composed. Therefore, how to improve the interpretability of deep metric learning methods is still an open problem that has barely been visited in previous works.

In this paper, we present a deep interpretable metric learning (DIML) framework as a first step towards more transparent embedding learning. Different from most existing deep metric learning methods that match the images by directly comparing feature vectors, we propose to leverage the spatial structure of images during matching to improve interpretability, as illustrated in Figure~\ref{fig:intro}. More specifically, we measure the similarity of two images by computing an optimal matching flow between the feature maps using the optimal transport theory such that the similarity can be decomposed into several part-wise similarities with different contributions to the overall similarity. Our framework consists of three key components: 1) \textbf{Structural Similarity (SS)}.  Unlike most existing deep metric learning methods that match the images by comparing feature vectors, we propose a new similarity/distance metric by measuring the similarity of corresponding parts in the feature maps based on the optimal matching flow; 2)  \textbf{Spatial Cross-Correlation (CC)}. To handle the view variance in the image retrieval problem, we propose to use spatial cross-correlation as the initial marginal distribution to compute the optimal transport plan; 3) \textbf{Multi-scale Matching (MM)}. We also devise a multi-scale matching strategy to better incorporate existing metric learning methods and enable us to adaptively adjust the extra computational cost in large-scale search problems. Since our method is model-agnostic and our contribution is orthogonal to previous deep metric learning methods on architectures~\cite{jacob2019horde}, objective functions~\cite{sohn2016improved_npair,kim2020proxy} and sampling strategies~\cite{wu2017sampling,zheng2019hardness}, our method can be applied to off-the-shelf backbone networks and metric learning methods even without training. Extensive experimental study on three major benchmarks of deep metric learning including CUB200-2011~\cite{cub}, Cars196~\cite{cars} and Stanford Online Products (SOP)~\cite{oh2016deep_sop} shows that our method enables us to achieve more interpretable metric learning while substantially improving various metric learning methods with or without re-training the models.


\section{Related Work}

\begin{figure*}
    \centering
    \includegraphics[width=\textwidth]{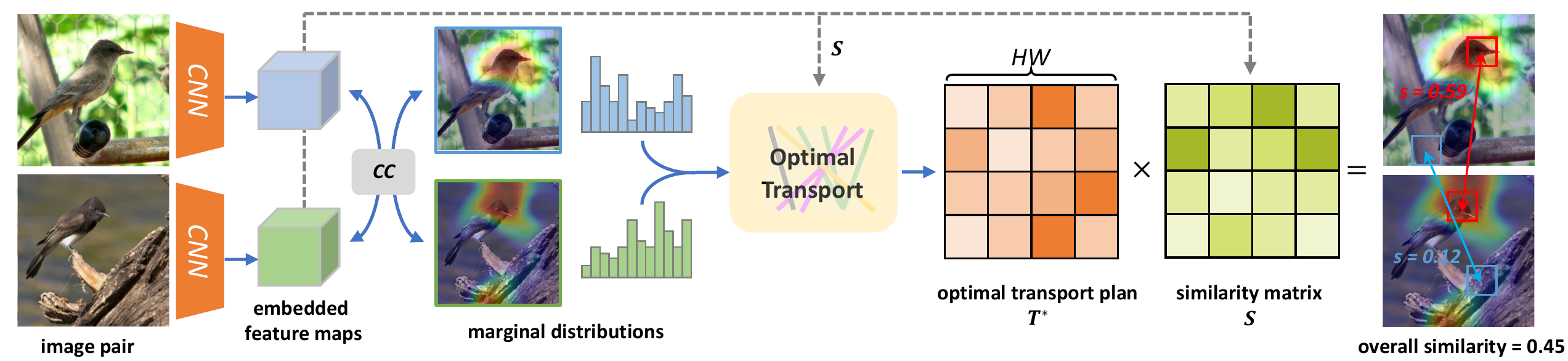}
    \caption{The overall pipeline of our deep interpretable metric learning (DIML) framework. The feature maps extracted from the backbone CNN model are further fed into the cross-correlation module (CC) to compute the marginal distributions that represent the weights of each location. The optimal transport plan then is obtained using the marginal distributions and the similarity matrix. Our framework decomposes the visual similarity to part-wise similarities and their contributions, which enable us to interpret and analyze how a deep model distinguishes two images.}
    \label{fig:framework}
\end{figure*}

\paragrapha{4pt}{Deep Metric Learning.} Deep metric learning (DML) has drawn increasing attention recently and become one of the primary framework for a range of vision tasks including image retrieval~\cite{sohn2016improved_npair, kim2020proxy}, image clustering~\cite{schroff2015facenet}, person re-identification~\cite{Beyond-triplet-loss-reid,rao2019learning,chen2020temporal} and face recognition~\cite{taigman2014deepface,deng2019arcface,rao2017attention}. Previous works on deep metric learning commonly focus on learning more accurate and robust embeddings to better reflect the semantic relations among samples. To achieve this goal, a variety of deep metric learning approaches are proposed to improve the architectures~\cite{xuan2018randomized-ensembles,jacob2019horde}, objective functions~\cite{hadsell2006dimensionality_contrastive,schroff2015facenet,chen2017beyond_quadruplet,oh2016deep_sop,sohn2016improved_npair,kim2020proxy} and sampling strategies~\cite{wu2017sampling,ge2018hierarchical-triplet,lin2018variational,zheng2019hardness,roth2020Policy-adapted-sampling}. 
Different from these works, there is a line of deep metric learning research on developing more effective distance or similarity metrics. 
Except for the commonly used $\ell^p$ distance and cosine similarity, signal-to-noise ratio (SNR)~\cite{yuan2019SNR} and hyperbolic geodesic distance~\cite{khrulkov2020hyperbolic} have also proven to be effective to reflect the semantic relationships among samples. However, these deep metric learning methods only consider the distance or similarity between feature vectors, which ignores the spatial structure of images and thus lacks interpretability. In this work, we propose to measure the similarity of two images by explicitly leveraging the spatial structures of images such that more accurate and interpretable similarity of two samples can be obtained.

\paragrapha{4pt}{Explainable \& Interpretable Vision Models.} Recent years have witnessed remarkable progress in various computer vision tasks driven by the success of deep learning~\cite{krizhevsky2012imagenet,he2016deep,lecun2015deep}. Despite the impressive discriminative power, the interpretability is often viewed as an Achilles’ heel of deep models. Improving the explainability and interpretability of deep models has attracted increasing attention in recent years. Existing works can be roughly divided into two groups: 1) explaining existing models through visualization and diagnosis of deep representations; 2) modifying deep models to learn disentangled and interpretable representations. For example, Zhou \etal~\cite{cam} proposes a method named Class Activation Mapping (CAM), which identifies discriminative regions in feature maps of CNNs by analyzing the effects on the final classification results. Grad-CAM~\cite{gradcam} improves the method by combining both the input features and the gradients of a model's layer. Apart from these methods focusing on explaining and analyzing trained models, interpretable vision models are developed by revising the architectures or training procedure of conventional deep models. Zhang \etal~\cite{zhang2018interpretable} design interpretable CNNs by enforcing each filter in a high-level convolutional layer represents a specific object part. \cite{zhang2019interpretingtree} combine the CNNs and decision tree to inherit the advantages of the two types of models to construct power yet interpretable image classification models. However, these methods only explain the reasoning process of how the output of a deep model is produced and did not consider the interaction between samples. Therefore, they cannot analyze and explain how the similarity of the two samples is composed. Recently, Williford \etal~\cite{williford2020explainable} present a study on explainable face recognition, which uses image editing techniques to generate a new dataset to evaluate what regions contribute to face matching. Their benchmark requires prior knowledge on face structures and thus is hard to generalize to other image matching problems. Different from these works, we propose to study a new and more generic problem of interpretable deep metric learning and provide a basic solution. 

\section{Approach}


\subsection{Preliminaries: Deep Metric Learning}

Deep metric learning aims to find a distance metric parameterized by deep neural networks to map the input image feature pairs to a distance in $\mathbb{R}$ that reflects the semantic similarity of the two images defined by labels. 
Formally, given a set of images $\mathcal{X} =\{x^k\}_{k=1}^N $ and the corresponding labels $\mathcal{Y} = \{y^k\}_{k=1}^N$, deep metric learning introduces the deep neural networks $f: \mathcal{X} \to \Phi\subset \mathbb{R}^C$ to map an image to a feature $\phi^k=f(x^k)$, where the semantic patterns of the input image are extracted. The mainstreams of deep metric learning aim to learn Mahalanobis distance metrics $d(\cdot, \cdot$), which can be formulated as:
\begin{equation}
\begin{split}
    d(x^k, x^l) = 	\| Mf(x^k) - Mf(x^l)\|_2 = \|g(\phi^k)- g(\phi^l)\|_2, \nonumber
\end{split}
\end{equation}
where $g(\phi^k)=M\phi^k:=\psi^k\in \Psi$ is an parametrized linear projection from the feature space $\Phi$ to an embedding space $\Psi\subset \mathbb{R}^D$. Following the configuration in the backbone networks like ResNet~\cite{he2016deep} and Inception~\cite{szegedy2015going}, $f$ can be decomposed into $f=\GAP \circ f_1$, where $f_1$ extracts a feature map $\omega^k=f_1(x^k)\in \mathbb{R}^{H\times W\times C}$ and $\GAP$ is the global average pooling. The $\GAP$ operation abstracts the feature maps into vectors so as to enable fast similarity calculation.



However, the abstraction on deep features also loses the spatial structures of the images during the embedding process, which makes most deep metric learning methods lack interpretability---deep models can tell us whether the two images are similar but cannot show us the reason. Since it is of importance to understand the matching mechanism in many risky visual applications, developing a more interpretable deep metric learning method becomes a critical research topic but it has barely been visited in previous works. 

\subsection{Structural Matching via Optimal Transport}\label{sec:optimal_transport}

To exploit the spatial structures in images for more interpretable deep metric learning, we devise a new structural matching scheme to compute feature similarity based on optimal transport theory~\cite{villani2008optimal}. 

Our core algorithm is adopted from the optimal transport theory, which aims to seek the \emph{minimal cost transport plan} between two distributions. Given a source distribution $\mu^\sss$ and a target distribution $\mu^\ttt$ that are defined on probability space $\mathcal{U}$ and $\mathcal{V}$ respectively, the minimal cost transport plan can be obtain by minimizing the Wasserstein distance between the two distributions:
\begin{equation}
    \pi^* = \arginf_{\pi \in \Pi(\mu^\sss, \mu^\ttt)} \int_{\mathcal{U} \times \mathcal{V}} c(u, v) {\rm d} \pi(u, v),\label{equ:wass}
\end{equation}
where $\pi^*$ is the optimal transport plan,  $\Pi(\mu^\sss, \mu^\ttt)$ is the joint probability distribution with marginals $\mu^\sss$ and  $\mu^\ttt$, and $c: \mathcal{U} \times \mathcal{V} \to \mathbb{R}^+$ is the cost function of transportation. 

Different from the above generic formulation, here we only need to consider the discrete distribution matching for image feature maps. Consider two feature maps $\omega^\sss,\omega^\ttt\in \mathbb{R}^{H\times W\times C}$ obtained by a backbone (\eg ResNet50~\cite{he2016deep}).  We first use the projection layer $g$ to map each element in the feature maps $\omega_{i}^k$ into an embedding space of dimension $D$ individually\footnote{For the sake of simplicity, we use a single subscript $i\in [1, HW]$ to index the spatial location. For pre-trained metric learning models, we can directly apply the original projection layer on the elements in the feature maps. Thus, our method does not need any modifications on the parameters. }:
\begin{equation} 
    z_{i}^\sss = g(\omega_{i}^\sss)\in \mathbb{R}^D,\quad z_{j}^\ttt = g(\omega_{j}^\ttt)\in \mathbb{R}^D.
\end{equation}
The cost of transporting one unit of mass from $i$ to $j$ is: 
\begin{equation}
    C_{i, j} =c(i, j):= d(z^\sss_i, z^\ttt_j),
\end{equation}
where we use the distance metric $d(\cdot, \cdot)$ for two vectors (\eg, Euclid distance or cosine distance) as the transport cost function $c$. In this discrete case, the transport plan $\pi$ matching the two distributions also becomes discrete. Given the two corresponding discrete distributions $\mu^\sss$ and $\mu^\ttt$, the original optimal transport problem is equivalent to:
\begin{equation}
\begin{split}
& T^* = \argmin_{T \geq 0} \text{tr}(CT^\top), \\
& \text{subject to} \quad T\mathbf{1} = \mu^\sss, \quad T^\top \mathbf{1} = \mu^\ttt.
\end{split}
\label{equ:optimal_transport}
\end{equation}
$T^*$ is the optimal matching flow between these two distributions, which can be also viewed as the structural matching plan of the two images. $T_{i,j}^*$ is the amount of mass that needs to move from $i$ to $j$ in order to reach an overall minimum cost, which represents the contribution of location pair $(i, j)$ to the overall matching. 

To efficiently solve the optimization problem in \eqref{equ:optimal_transport}, we adopt the Sinkhorn divergence algorithm~\cite{cuturi2013sinkhorn} by introducing an entropic regularizer to enable fast training and inference. More details about the algorithm can be found in Supplementary Material. 
Note that the iterative algorithm is fully differentiable, which can be easily implemented by using the automatic differentiation library like PyTorch~\cite{paszke2019pytorch} and directly apply the matching process to any deep metric learning pipelines.

\paragrapha{4pt}{Discussions.}  Some closely related works of the proposed structural matching scheme include EMD metric learning~\cite{zhang2018emd-Metric-Learning} and Wasserstein embedding learning~\cite{frogner2019learning-entropic-wasserstein-spaces}. However, different from our method, they usually focus on learning better embeddings for set inputs, which can be naturally solved by the Wasserstein distance metric learning framework. Here our main contribution is not the matching algorithm itself but the introduction of structural matching for learning more interpretable visual similarity.

\subsection{Deep Interpretable Metric Learning}~\label{sec:diml}
In Section~\ref{sec:optimal_transport}, we have already shown how to calculate the distance between two distributions using the optimal transport. In this section, we describe how to perform structural matching in metric learning. Specifically, our method consists of three components: 1) we use optimal transport to calculate the \textit{structural similarity} (SS) of two images; 2) we propose to calculate the spatial \textit{cross-correlation} (CC) to initialize the marginal distributions in Equation~\eqref{equ:wass}; 3) we propose \textit{multi-scale matching} (MM) to improve the metric and reduce the computation cost.

\paragrapha{4pt}{Structural Similarity (SS).} Given the marginal distributions $\mu^\sss$ and $\mu^\ttt$ (which we will discuss in detail later) and the cost matrix $C$, we can then obtain the optimal transport $T^*$ by solving~\eqref{equ:optimal_transport}. Once we have $T^*$, we can define the structural similarity of two feature maps $z^\sss,z^\ttt\in \mathbb{R}^{HW\times D}$ as follows:
\begin{equation}
    s_\structural{} (z^\sss, z^\ttt)=\sum_{1\le i,j\le HW}s(z_i^\sss, z_j^\ttt)T^*_{i,j},\label{equ:diml_sim}
\end{equation}
where $s(\cdot, \cdot)$ is a function to calculate the similarity between two vectors. Our structural similarity enables us to investigate the composition of the overall similarity, thus we can easily decompose the similarity and understand how the similarity between different locations in the two images contribute to the overall similarity. Similarly, given any distance function $d(\cdot, \cdot)$, we can also derive our structural distance:


\begin{equation}
        d_\structural{}(z^\sss, z^\ttt)=\sum_{1\le i,j\le HW}d(z_i^\sss, z_j^\ttt)T^*_{i,j},\label{equ:diml_distance}
\end{equation}

\paragrapha{4pt}{Cross-Correlation (CC).} Another important part is the definition of the marginal distributions $\mu^\sss$ and $\mu^\ttt$. One trivial solution is to initialize them with uniform distributions, \ie, 
\begin{equation}
    \mu^\sss_i=\mu^\ttt_i=\frac{1}{HW}, \forall 1\le i\le HW,
    \label{equ:uniform}
\end{equation}
which indicates similarity of each location has the identical weight to the overall similarity. 
In the structural matching algorithm, the marginal distributions should characterize the importance of each spatial location. Simply using uniform distributions implies that we want to match all the features with equal importance, which is not desired in some cases. For example, some image contains background information that may be less useful for matching thus we want to impose lower weights on the background. Another common circumstance is when we want to match two images with different views (\eg, the first image contains the whole object and the second one only contains a part of it), and similarly we only need to focus on the certain part of the first image and treat the rest as background. To find the areas that are most related to the similarity, we propose to calculate the cross-correlation between the two images as the marginal distributions for the matching algorithm. Specifically, we first perform global average pooling to $z^\sss,z^\ttt$ and obtain the global feature $\bar{z}^\sss,\bar{z}^\ttt$. We then slide the global feature of one image on the feature map of the other image and calculate point-wise correlation at each spatial location. Formally, the cross-correlation is calculated as:
\begin{equation}
    \alpha_i^\sss = \frac{\langle \bar{z}^\sss, z^\ttt_i \rangle}{\|\bar{z}^\sss\|\|z^\ttt_i\|},\alpha_i^\ttt = \frac{\langle \bar{z}^\ttt, z^\sss_i \rangle}{\|\bar{z}^\ttt\|\|z^\sss_i\|},
\end{equation}
where $\langle\cdot, \cdot, \rangle$ is the dot product and $\alpha_i^{k}\in [-1, 1]$. After obtaining the cross-correlation, we can use $\alpha_i^k$ to reflect the importance of $z^{k}_i$ in the matching problem. To further reduce the effects of low correlation regions, we discard the negative value of $\alpha_i^{k}$ and normalize it to obtain the final marginal distributions:
\begin{equation}
    \begin{aligned}
    &\gamma_i^k= \max(0, \alpha_i^k), \mu^k_i= \frac{1}{\sum_{i'}\gamma^k_{i'}}\gamma_i^k\\
    &\forall 1\le i\le HW, \quad k\in \{\sss, \ttt\}.
\end{aligned}
\end{equation}
Once we have the marginal distributions $\mu^{(k)}$, we can then apply the structural matching algorithm in Section~\ref{sec:optimal_transport} to calculate the similarity between two images. We will show in Section~\ref{sec:ablation} that cross-correlation is an indispensable component to improve the power of DIML.

\paragrapha{4pt}{Multi-scale Matching (MM).} Although DIML can capture the structural similarity of two images and can provide results easily understood by humans, it requires more computation ($\mathcal{O}(H^2W^2)$) to solve the optimal transport problem. In the application of image retrieval, there are usually a great number of images in the gallery. Given an image as an anchor, calculate the structural similarity between the anchor and all the images in the gallery is inefficient. To reduce the computational cost, we propose a multi-scale matching method for image retrieval. Let $z^{\rm a}\in \mathbb{R}^{H\times W\times D}$ be the feature map of the anchor image and $z^{k}\in \mathbb{R}^{H\times W\times D}, k=1,\ldots, N$ be the feature maps of all the images in the gallery. In the first scale ($1\times 1$), we compute the global feature using global average pooling to get $\bar{z}^{\rm a}, \bar{z}^{k}\in\mathbb{R}^D$, and calculate cosine similarity between $\bar{z}^a$ and each $\bar{z}^{k}$ as conventional DML methods. We can then define a truncation number $K$ and select the images with top-$K$ similarity score and denote the indices of them as $\mathcal{I}_K$ to further enhance the similarity with our method. In the second scale ($H\times W$), we calculate the structural similarity between $z^a$ and each $z^{k}, k\in \mathcal{I}_K$. Since $K$ is fixed, the extra computational cost of DIML can be controlled. By multi-scale matching, we can filter out the obvious dissimilar samples ($1\times 1$ scale, cosine similarity) and focus on the hard ones ($H\times W$ scale, structural similarity). Combining the similarity at two scales can also capture both semantic and spatial information, which is also helpful to improve retrieval precision. We will show later in Section~\ref{sec:ablation} that a small $K$ can yield a significant performance boost.

\subsection{Implementation}\label{sec:implementation}
One of the major advantages of DIML is that we can apply DIML to any pre-train model to improve performance with \textbf{\textit{no need of training}}. Besides, we can also incorporate DIML into the training objective. In this section, we will describe how to use DIML in these two scenarios.

\paragrapha{4pt}{Testing.} Given a pre-trained model, we first calculate the feature maps $\omega^{\sss}, \omega^{\ttt}\in\mathbb{R}^{H\times W\times C}$ (before the global pooling layer) of the image pair $x^\sss, x^\ttt$. We can then use the algorithm describe in Section~\ref{sec:diml}~to compute the structural similarity. However, $HW$ may be sometimes large in practice (\eg, for ResNet50~\cite{he2016deep},  $H=W=7$). Therefore, we can use ROI Align~\cite{he2017mask} to pool the feature maps to $\mathbb{R}^{H'\times W'\times C}$, where $H' < H$ and $W' < W$. With smaller feature maps, we can then calculate the structural similarity with a relatively lower computational cost. In our implementation, we use $H'=W'=G$ and we found $G=4$ can achieve good trade-off between cost and performance.

\paragrapha{4pt}{Training.} We can also combining DIML and existing metric learning methods to facilitate training. We now use Margin loss~\cite{wu2017sampling} as an example to show how to incorporate DIML into the training objective. The Margin loss~\cite{wu2017sampling} is defined as
\begin{equation}
    \mathcal{L}_{\rm margin}(k, l) =\left(\sigma+(-1)^{I(y_k\neq y_l)}\left(D_{k,l}-\beta\right)\right)_{+},
\end{equation}
where $\sigma$ and $\beta$ are learnable parameters, and $D_{kl}$ is used to measure the distance between image $k$ and $l$:
\begin{equation}
    D_{k,l} = \frac{1}{2}\left(d_\structural{}(z^k, z^l) + d(\bar{z}^k, \bar{z}^l)\right)
\end{equation}
For the implementation details of other metric learning methods, please refer to the Supplementary Material.


\section{Experiments}

To evaluate the performance of our proposed \ours, we conduct experiments on three widely used datasets in the image retrieval research field: CUB200-2011~\cite{cub}, Cars196~\cite{cars}, and Standard Online Products (SOP)~\cite{oh2016deep_sop}. 

\subsection{Experiment Setups}

\paragrapha{4pt}{Datasets.} We evaluate our method under a zero-shot image retrieval setting, where the
training set and test set contain image classes with no intersection. We follow the training/test set splits in previous works~\cite{musgrave2020metric,roth2020revisiting}: 
\begin{itemize}
    \item  CUB200-2011~\cite{cub} contains 11,788 images of birds from 200 species. The first 100 classes (5,864 images) are used for training, while other 100 classes (5,924 images) are kept for testing. 
    \item  Cars196~\cite{cars} contains 16,185 images of cars from 196 classes. We use the first 98 classes (8,054 images) for training and leave the rest 98 classes (8,131 images) for testing. 
    \item  SOP~\cite{oh2016deep_sop} contains 120,053 images from 22,634 classes. We use the first 11,318 classes (59,551 images) for training and other 11,316 (60,502 images) for testing.
\end{itemize} 

\paragrapha{4pt}{A Fair Evaluation Protocol.} Although there are many previous metric learning methods, \cite{musgrave2020metric} pointed out that the improvements over time are not significant, due to the unfair comparisons of different methods. Therefore, we try our best to provide fair results by implementing all the methods under the same evaluation protocol. For all the baseline methods, we use ResNet-50~\cite{he2016deep} pre-trained on ImageNet~\cite{krizhevsky2012imagenet} as the backbone. We freeze the BatchNorm layers during training and modify the output channel of the last linear layer to a fixed embedding dimension $D$. We use embedding size $D=128$ and other implementation settings following~\cite{roth2020revisiting} for most experiments unless otherwise noted.

\paragrapha{4pt}{Evaluation Metrics.} 
Most previous works use Recall@K, Normalized Mutual Information, and F1 score as accuracy metrics. However, as is pointed by~\cite{musgrave2020metric}, NMI and F1 scores sometimes give us wrong pictures of the embedding space. To this end, we adopt the evaluation metrics used in~\cite{musgrave2020metric}: Precision@1, R-Precision, and MAP@R. For the formal definition of the metrics, see Supplementary Material.

\paragrapha{4pt}{Implementation.} It is also worth noting that our proposed \ours~does not require any training. Therefore, we aim to prove that our method can improve the performance given \textit{any} trained model as the baseline. Therefore, we perform experiments on a wide range of loss functions
(Margin~\cite{wu2017sampling}, Arcface~\cite{deng2019arcface}, \etc) 
and sampling methods 
(Distance~\cite{wu2017sampling}, N-Pair~\cite{sohn2016improved_npair}, \etc) 
to prove the effectiveness and the generalization ability of our method. For most of the baseline methods, we follow the implementation from~\cite{roth2020revisiting}. 





\begin{table}[t]
  \centering
  \caption{\textbf{Applying DIML to various deep metric learning methods.} Experimental results show that our method can improve the performance of all the methods consistently.} \vspace{-5pt}
 \adjustbox{max width=0.98\linewidth}{
 \setlength{\tabcolsep}{1pt}
   \begin{tabu}to 1.45\linewidth{l*{3}{X[c]}|*{3}{X[c]}|*{3}{X[c]} } \toprule
    \multicolumn{1}{l}{\multirow{2}[0]{*}{\textbf{Method}}} & \multicolumn{3}{c}{\textbf{CUB200-2011}} & \multicolumn{3}{c}{\textbf{Cars196}} & \multicolumn{3}{c}{\textbf{SOP}} \\\cmidrule(lr){2-4}\cmidrule(lr){5-7}\cmidrule(lr){8-10}
          & \multicolumn{1}{c}{\textbf{P@1}} & \textbf{RP} & \multicolumn{1}{c}{ \textbf{M@R}} & \textbf{P@1} & \textbf{RP} & \multicolumn{1}{c}{\textbf{M@R}} & \textbf{P@1} & \textbf{RP} & \textbf{M@R} \\\midrule

Contrasitive~\cite{hadsell2006dimensionality_contrastive} & 61.77 & 34.25 & 23.24 & 67.45 & 30.01 & 18.61 & 73.27 & 40.92 & 37.5 \\
+ \ours & \textbf{64.43} & \textbf{35.16} & \textbf{24.29} & \textbf{73.35} & \textbf{30.76} & \textbf{20.13} & \textbf{74.47} & \textbf{41.58} & \textbf{38.29} \\\midrule
Triplet-R~\cite{schroff2015facenet} & 58.34 & 32.00 & 20.93 & 62.73 & 26.95 & 15.24 & 66.60 & 33.62 & 30.26 \\
+ \ours & \textbf{60.60} & \textbf{32.63} & \textbf{21.74} & \textbf{67.92} & \textbf{27.65} & \textbf{16.72} & \textbf{68.73} & \textbf{35.04} & \textbf{31.79} \\\midrule
Triplet-S~\cite{schroff2015facenet} & 59.28 & 32.77 & 21.79 & 67.00 & 30.0 & 18.23 & 73.67 & 40.45 & 37.14 \\
+ \ours & \textbf{62.85} & \textbf{33.87} & \textbf{23.04} & \textbf{72.06} & \textbf{30.89} & \textbf{20.04} & \textbf{75.14} & \textbf{41.68} & \textbf{38.42} \\\midrule
Triplet-H~\cite{roth2020revisiting} & 61.39 & 33.21 & 22.15 & 71.76 & 32.53 & 20.83 & 73.28 & 39.98 & 36.56 \\
+ \ours & \textbf{62.02} & \textbf{33.50} & \textbf{22.49} & \textbf{74.75} & \textbf{32.94} & \textbf{21.76} & \textbf{73.62} & \textbf{40.14} & \textbf{36.79} \\\midrule
Triplet-D ~\cite{wu2017sampling} & 61.99 & 33.92 & 22.90 & 73.07 & 32.18 & 20.81 & 77.34 & 44.25 & 40.80 \\
+ \ours & \textbf{63.40} & \textbf{34.49} & \textbf{23.59} & \textbf{77.31} & \textbf{33.05} & \textbf{22.61} & \textbf{77.81} & \textbf{44.82} & \textbf{41.39} \\\midrule
NPair~\cite{sohn2016improved_npair} & 60.30 & 33.53 & 22.27 & 69.52 & 32.24 & 20.25 & 76.47 & 43.48 & 39.94 \\
+ \ours & \textbf{62.17} & \textbf{34.02} & \textbf{22.85} & \textbf{74.65} & \textbf{32.91} & \textbf{21.67} & \textbf{76.86} & \textbf{43.87} & \textbf{40.38} \\\midrule
Angular~\cite{wang2017deep} & 61.36 & 34.17 & 23.00 & 70.93 & 32.97 & 21.31 & 73.79 & 41.42 & 37.90 \\
+ \ours & \textbf{63.77} & \textbf{35.09} & \textbf{24.06} & \textbf{74.72} & \textbf{33.80} & \textbf{22.83} & \textbf{74.91} & \textbf{42.17} & \textbf{38.73} \\\midrule
GenLifted~\cite{hermans2017defense_genlift} & 58.27 & 32.86 & 21.83 & 66.88 & 30.96 & 19.00 & 74.84 & 42.28 & 38.66 \\
+ \ours & \textbf{61.07} & \textbf{33.82} & \textbf{22.98} & \textbf{72.95} & \textbf{31.93} & \textbf{20.88} & \textbf{75.92} & \textbf{43.08} & \textbf{39.55} \\\midrule
ProxyNCA~\cite{movshovitz2017no_pnca} & 62.76 & 35.05 & 24.03 & 71.05 & 31.62 & 20.55 & 74.70 & 41.32 & 37.96 \\
+ \ours & \textbf{64.75} & \textbf{36.02} & \textbf{25.10} & \textbf{74.86} & \textbf{32.43} & \textbf{22.00} & \textbf{76.17} & \textbf{42.65} & \textbf{39.36} \\\midrule
Histogram~\cite{evgeniya2016histogram} & 59.96 & 33.07 & 22.15 & 69.49 & 31.62 & 19.76 & 71.15 & 38.06 & 34.70 \\
 + \ours & \textbf{62.69} & \textbf{33.80} & \textbf{23.00} & \textbf{74.50} & \textbf{32.36} & \textbf{21.26} & \textbf{72.06} & \textbf{38.57} & \textbf{35.30} \\\midrule
Quadruplet~\cite{chen2017beyond_quadruplet} & 61.53 & 34.05 & 22.93 & 69.64 & 31.40 & 19.67 & 77.02 & 44.27 & 40.88 \\
+ \ours & \textbf{62.80} & \textbf{34.65} & \textbf{23.62} & \textbf{75.66} & \textbf{32.35} & \textbf{21.69} & \textbf{78.08} & \textbf{45.16} & \textbf{41.79} \\\midrule
SNR~\cite{yuan2019SNR} & 62.00 & 34.72 & 23.59 & 72.95 & 32.72 & 21.28 & 77.82 & 44.98 & 41.51 \\
+ \ours & \textbf{64.55} & \textbf{35.25} & \textbf{24.27} & \textbf{77.57} & \textbf{33.54} & \textbf{23.02} & \textbf{78.50} & \textbf{45.65} & \textbf{42.24} \\\midrule
Softmax~\cite{zhai2019classification_softmax} & 61.06 & 32.7 & 21.55 & 72.61 & 31.17 & 19.88 & 77.02 & 43.47 & 40.25 \\
+ \ours & \textbf{63.30} & \textbf{33.71} & \textbf{22.64} & \textbf{76.39} & \textbf{32.06} & \textbf{21.49} & \textbf{78.17} & \textbf{44.62} & \textbf{41.43} \\\midrule

Margin~\cite{wu2017sampling} & 62.47 & 34.12 & 23.14 & 72.18 & 32.00 & 20.82 & 78.39 & 45.64 & 42.34 \\ 
+ \ours & \textbf{65.16} & \textbf{35.37} & \textbf{24.51} & \textbf{76.62} & \textbf{32.85} & \textbf{22.48} & \textbf{79.26} & \textbf{46.44} & \textbf{43.19} \\\midrule
Arcface~\cite{deng2019arcface} & 61.39 & 33.70 & 22.4 & 73.37 & 31.90 & 20.52 & 77.55 & 44.44 & 41.07 \\
+ \ours & \textbf{64.72} & \textbf{34.88} & \textbf{23.72} & \textbf{77.24} & \textbf{32.88} & \textbf{22.34} & \textbf{78.52} & \textbf{45.45} & \textbf{42.10} \\\midrule
MS~\cite{wang2019multi} & 62.56 & 32.74 & 21.99 & 74.81 & 32.72 & 21.60 & 77.90 & 44.97 & 41.54 \\
+ \ours & \textbf{64.89} & \textbf{33.99} & \textbf{23.34} & \textbf{78.44} & \textbf{33.57} & \textbf{23.31} & \textbf{78.53} & \textbf{45.59} & \textbf{42.22} \\\midrule
ProxyAnchor~\cite{kim2020proxy} & 65.24 & 35.81 & 24.87 & 82.36 & 36.00 & 25.85 & 79.10 & 46.31 & 42.91 \\
+ \ours & \textbf{66.46} & \textbf{36.49} & \textbf{25.58} & \textbf{86.13} & \textbf{37.90} & \textbf{28.11} & \textbf{79.22} & \textbf{46.43} & \textbf{43.04} \\\bottomrule
    \end{tabu}
} \vspace{-10pt}
  \label{tab:main}
\end{table}

\subsection{Main Results}
We first evaluate our method by applying \ours~to a wide range of metric learning methods. We measure the performance using the evaluation metrics aforementioned: Precision@1 (P@1),  R-Precision (RP) and MAP@R (M@R), and the results\footnote{For more results, please refer to the Supplementary Material.} are shown in Table~\ref{tab:main}. For all the experiments, we set the truncation number $K=100$ and feature map size $G=4$. We observe that our method can improve the performance for \textbf{\textit{all}} the models on \textbf{\textit{all}} the three benchmarks, without any extra training. Especially, we find on Cars196 dataset, the performances of all the methods are enhanced profoundly after equipped with our DIML.


\subsection{Ablation Study and Analysis}
In this section, we will evaluate our \ours~in various settings and provide detailed analyses through experiments and visualization.\label{sec:ablation}

\paragrapha{0pt}{Effects of different components.} The \ours~consists of three components: structural similarity (SS), cross-correlation (CC), and multi-scale matching (MM). We will analyze the effect of each one, as is shown in Table~\ref{tab:ablation}. We start with two baseline methods Margin~\cite{wu2017sampling} and Multi-Similarity~\cite{wang2019multi}, and add the three components gradually. First, we adopt structural similarity instead of standard cosine similarity (where we use uniform distribution in Equation~\eqref{equ:uniform} for $\mu^\sss$ and $\mu^\ttt$). We find that SS can improve the performance on all the datasets \textit{except} for SOP (as is highlighted by underline). It is mainly because that the SS algorithm aims to match every part of a source image to a target image. However, the views vary a lot in SOP dataset, which hinders the application of SS. Second, we show that multi-scale matching can make use of the semantic information and improve the performance on all three datasets. Finally, we replace the uniform distribution with the one calculated by cross-correlation. We find the marginal distributions obtained in this way are helpful to explore the important area of the images and can further improve the performance.

\begin{table}[tbp]
  \centering
  \caption{\textbf{Effects of the three components in our \ours:} Structural Similarity (SS), Multi-scale Matching (MM) and Cross Correlation (CC). We show that our method can enhance the performance of the baseline methods by combining the three components together.}
  \adjustbox{max width=\linewidth}{
  \setlength{\tabcolsep}{2pt}
    \begin{tabu}to 1.45\linewidth{c*{3}{X[c]}*{6}{X[c]}}\toprule
    \multirow{2}[0]{*}{\textbf{Baseline}} & \multicolumn{3}{c}{\textbf{Components}} & \multicolumn{2}{c}{\textbf{CUB200-2011}} & \multicolumn{2}{c}{\textbf{Cars196}} & \multicolumn{2}{c}{\textbf{SOP}} \\\cmidrule(rl){2-4}\cmidrule(rl){5-6}\cmidrule(rl){7-8}\cmidrule(rl){9-10}
          & SS     & MM     & CC    & \textbf{P@1} &  \textbf{M@R} & \textbf{P@1} & \textbf{M@R} & \textbf{P@1}  & \textbf{M@R} \\\midrule
    \multirow{4}[0]{*}{Margin~\cite{wu2017sampling}} &       &       &       & 62.47   & 23.14  & 72.18    & 20.82  & 78.39   & 42.34  \\
          & \cmark &       &       & 63.64    & 22.52  & 74.86    & 21.24  & \textit{\ul{77.30}}  &  \textit{\ul{41.02}}  \\
          & \cmark & \cmark &       & 64.96    & 23.87  & 76.02    & 22.02  & 78.53    & 42.45  \\
          & \cmark & \cmark & \cmark & \textbf{65.16} & \textbf{24.51} & \textbf{76.62}  & \textbf{22.48} & \textbf{79.26} & \textbf{43.19} \\\midrule
    \multirow{4}[0]{*}{MS~\cite{wang2019multi}} &       &       &       & 62.56    & 21.99  & 74.81   & 21.60  & 77.90 & 41.54  \\
          & \cmark &       &       & 63.52   & 21.66  & 75.63    & 21.10  & \textit{\ul{75.81}}    & \textit{\ul{39.38}}  \\
          & \cmark & \cmark &       & 64.40    & 22.83  & 77.39   & 22.77  & 77.87   & 41.55  \\
          & \cmark & \cmark & \cmark & \textbf{64.89} & \textbf{23.34} & \textbf{78.44} & \textbf{23.31} & \textbf{78.53}  & \textbf{42.22} \\\bottomrule
    \end{tabu}%
    }
  \label{tab:ablation}%
\end{table}%
\begin{figure}
    \centering
    \includegraphics[width=\linewidth]{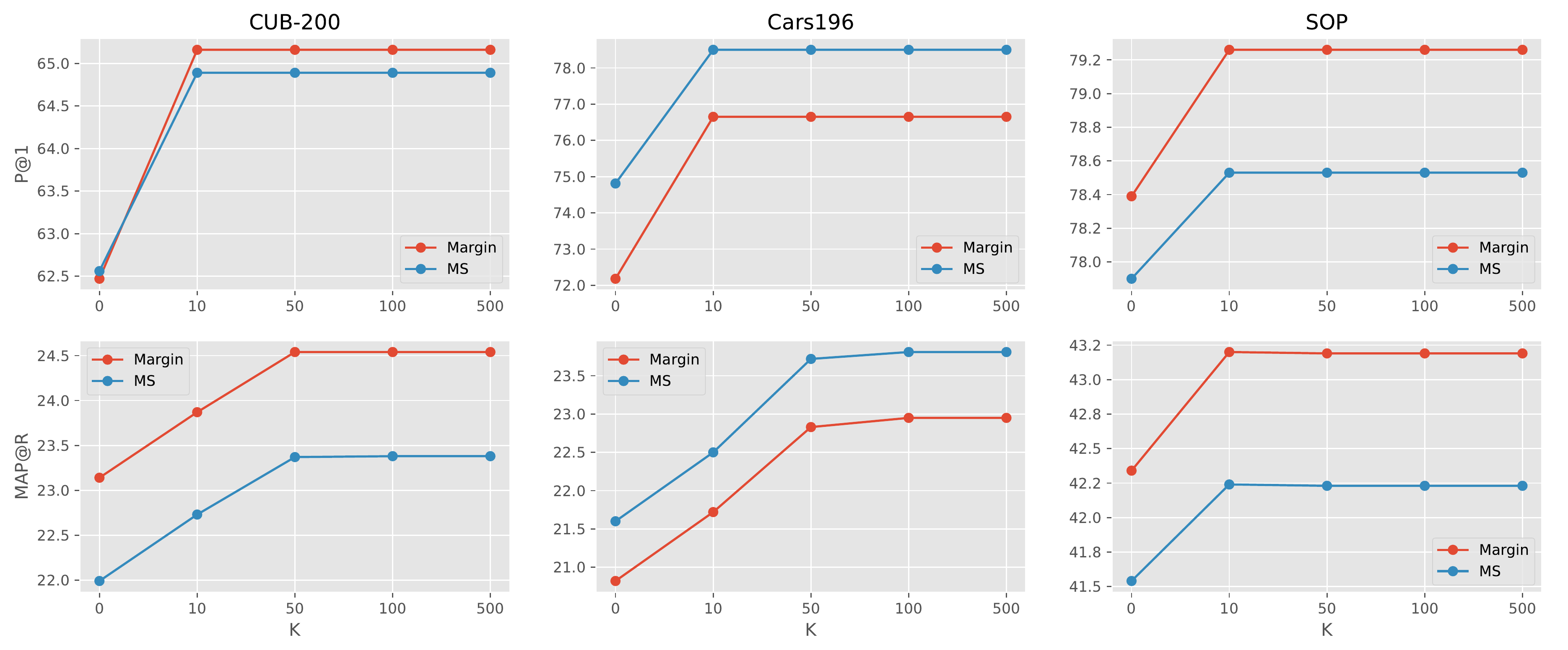}
  \caption{\textbf{Comparisons of different truncation number.} We test for different truncation number $K$ ranging from 0 to 500. Experimental results show that a small $K$ can already bring considerable performance improvement.}
    \label{fig:truncation_number}
\end{figure}

\paragrapha{4pt}{Effects of truncation number.} To show how the truncation number $K$ affect our DIML, we test our method on Margin~\cite{wu2017sampling} and Multi-Similarity~\cite{wang2019multi} with $K$ increasing from 0 to 500 (Figure~\ref{fig:truncation_number}). Note that $K=0$ means no structural similarity is used, which is identical to the baseline. We find that even a small $K$ will bring considerable improvement on the performance (especially for the P@1 metric). Generally, the retrieval accuracy grows with $K$ increasing and saturates before $K$ reaches 100. This phenomenon indicates that with a fixed and relatively small $K$, we can already enjoy a significant performance boost with constant extra computational cost and no extra training cost.

\begin{figure}
    \centering
    \includegraphics[width=\linewidth]{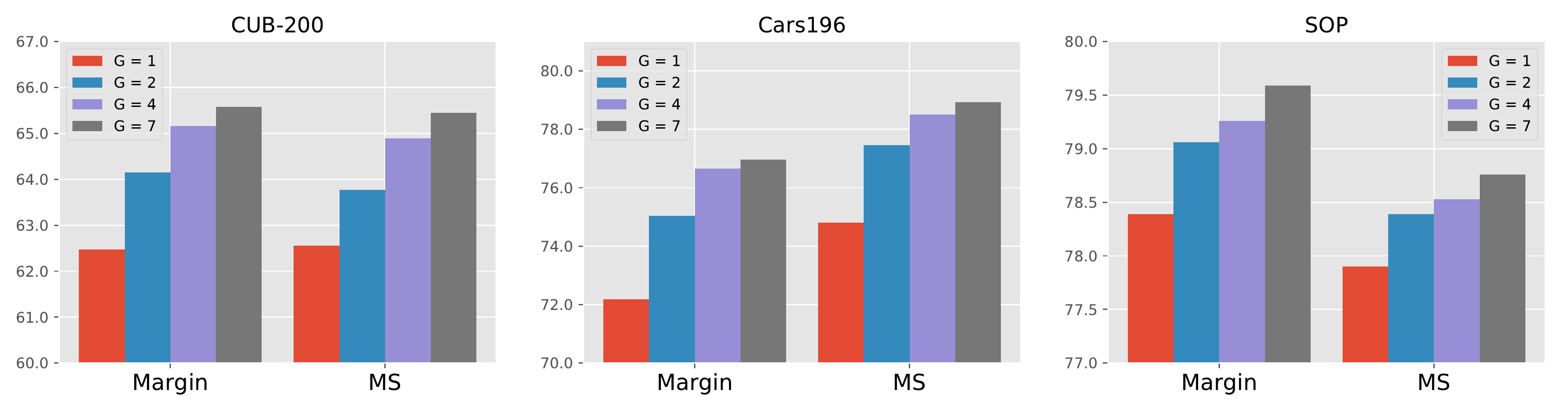} \vspace{-5pt}
  \caption{\textbf{Effects of the size of feature map.} Generally, the performance of our \ours~is better with higher $G$. \ours~with $G=4$ yields good results within relatively low computational costs.}\label{fig:fm_size}
\end{figure}
\paragrapha{4pt}{Effects of feature map size.} We then perform an ablation study on the feature map size $G$. In our experiments, we use ResNet50~\cite{he2016deep} as our backbone, and the size of the feature map before the pooling layer is $7\times 7$. Hence, we need to pool the feature map to a smaller one ($G\times G$) to reduce computational complexity. Specifically, we let $G \le  7$ and evaluate for the cases where $G=1,2,4,7$. The results are shown in Table~\ref{fig:fm_size} and we observe that the performance of our method is better with larger $G$ in general. We can also find $G=4$ is a good trade-off between performance and computational complexity.

\begin{table}[tbp]
  \centering
  \caption{\textbf{Effects of training.} Our method can substantially improve the baseline model with or without training.}
  \label{tab:training}%
  \vspace{-5pt}
  \adjustbox{max width=\linewidth}{
    \begin{tabular}{crccccccc}\toprule
    \multirow{2}[0]{*}{\textbf{Baseline}} & \multicolumn{2}{c}{\textbf{Setting}} & \multicolumn{2}{c}{\textbf{CUB200-2011}} & \multicolumn{2}{c}{\textbf{Cars196}} & \multicolumn{2}{c}{\textbf{SOP}} \\\cmidrule(lr){2-3}\cmidrule(lr){4-5}\cmidrule(lr){6-7}\cmidrule(lr){8-9}
          & \multicolumn{1}{l}{\textbf{test}} & \multicolumn{1}{l}{\textbf{train}} & \textbf{P@1}  & \textbf{M@R} & \textbf{P@1}  & \textbf{M@R} & \textbf{P@1}  & \textbf{M@R} \\\midrule
    \multirow{3}[0]{*}{Margin~\cite{wu2017sampling}} &       &       & 62.47  & 23.14  & 72.18   & 20.82  & 78.39   & 42.34  \\
          & \cmark     &       & 65.16   & 24.54  & \textbf{76.65}   & \textbf{22.95}  & \textbf{79.26}    & \textbf{43.19}  \\
          & \cmark     & \cmark     & \textbf{65.36}    & \textbf{24.90}  & 75.61    & 22.34  & 78.81    & 42.89  \\\midrule
    \multirow{3}[0]{*}{MS~\cite{wang2019multi}} &       &       & 62.56    & 21.99  & 74.81    & 21.60  & 77.90    & 41.54  \\
          & \cmark     &       & 64.89  & 23.38  & 78.50    & \textbf{23.81}  & 78.53   & 42.23  \\
          & \cmark     & \cmark     & \textbf{65.72}  & \textbf{24.37}  & \textbf{78.90}    & 23.80  & \textbf{79.00}   & \textbf{42.96}  \\\bottomrule
    \end{tabular}%
    }
\end{table}%

\paragrapha{4pt}{Effects of training.} Besides the default setting where we use \ours~to test on any pre-trained model, we can also incorporate the structural similarity into the training objectives (see Section~\ref{sec:implementation} for details). In Table~\ref{tab:training}, we compare the performance in three scenarios: (1) without DIML testing or training (same as baseline) (2) with DIML testing only (2) with DIML testing and training. The results are listed in Table~\ref{tab:training}. We find that for most cases, using DIML to test a pre-trained model can already improve the baseline by a significant margin. Besides, it is also useful sometimes to apply DIML in the training stage.

\begin{figure*}
    \centering
    \includegraphics[width=0.94\textwidth]{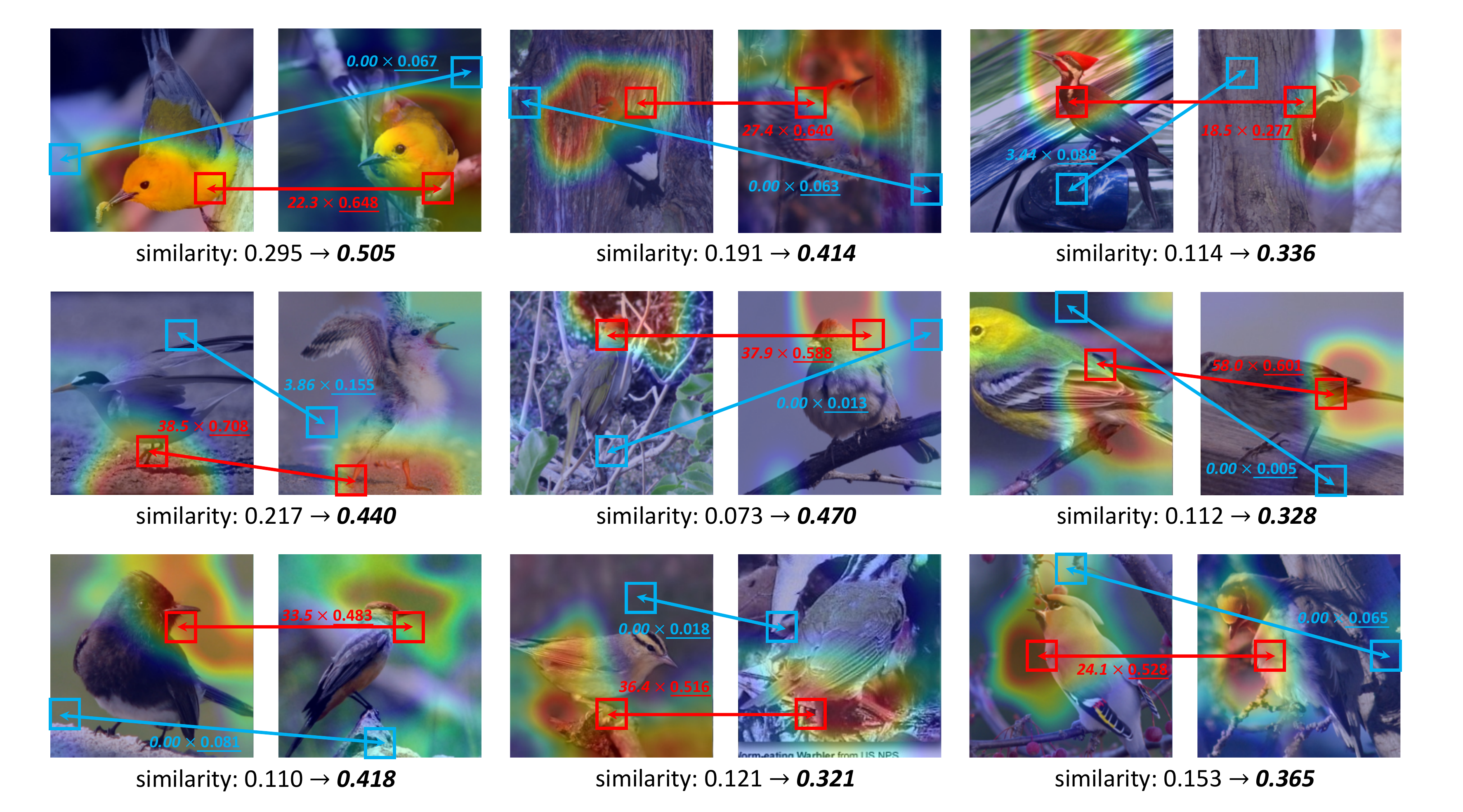}
    \caption{\textbf{Visualization}. We use heatmaps to show the marginal distributions obtained by cross-correlation (CC). We also illustrate two most representative part-wise similarity and their contributions to the overall similarity in the form of $[G^4 T^*_{i,j}]\times [\underline{S_{i,j}}]$, where $G$ is the grid size, $T^*_{i,j}$ and $\underline{S_{i,j}}$ are the matching flow and similarity between location $i$ and $j$ respectively. We also show the overall similarity changes after applying our method to the baseline model (cosine similarity $\to$ \textbf{\emph{structural similarity}}). All image pairs are positive pairs.  } \vspace{-10pt}
    \label{fig:vis}
\end{figure*}

\begin{table}[tbp]
  \centering
\caption{\textbf{Effects of embedding size.} Our~\ours~is robust to the changing of the embedding size $D$ and can improve the performance of the baseline methods consistently.} \vspace{-5pt}
  \adjustbox{max width=\linewidth}{
    \begin{tabular}{clcccccc}\toprule
    \multirow{2}[0]{*}{$D$} & \multicolumn{1}{c}{\multirow{2}[0]{*}{\textbf{Method}}} & \multicolumn{3}{c}{\textbf{CUB200-2011}} & \multicolumn{3}{c}{\textbf{Cars196}} \\\cmidrule(lr){3-5}\cmidrule(lr){6-8}
          &       & \textbf{P@1} & \textbf{RP} & \textbf{M@R} & \textbf{P@1} & \textbf{RP} & \textbf{M@R} \\\midrule
    \multirow{6}[0]{*}{64} & Margin & 59.39  & 32.59  & 21.53  & 69.31  & 30.98  & 19.67  \\
          & Margin~\cite{wu2017sampling} + \ours & \textbf{62.98}  & \textbf{33.88}  & \textbf{22.95}  & \textbf{74.44}  & \textbf{31.96}  & \textbf{21.60}  \\\cmidrule{2-8}
          & MS~\cite{wang2019multi} & 58.52  & 31.35  & 20.23  & 71.67  & 30.90  & 19.57  \\
          & MS + \ours & \textbf{61.73}  & \textbf{32.61}  & \textbf{21.62}  & \textbf{76.94}  & \textbf{31.95}  & \textbf{21.72}  \\\cmidrule{2-8}
          & ProxyAnchor~\cite{kim2020proxy} & 62.56  & 34.61  & 23.45  & 78.08  & 34.35  & 23.95  \\
          & ProxyAnchor + \ours & \textbf{65.01}  & \textbf{35.53}  & \textbf{24.40}  & \textbf{83.11}  & \textbf{36.49}  & \textbf{26.55}  \\\midrule
    \multirow{6}[0]{*}{512} & Margin~\cite{wu2017sampling} & 64.92  & 35.94  & 24.92  & 73.68  & 32.03  & 21.09  \\
          & Margin + \ours & \textbf{66.91}  & \textbf{36.82}  & \textbf{25.89}  & \textbf{76.67}  & \textbf{32.62}  & \textbf{22.20}  \\\cmidrule{2-8}
          & MS~\cite{wang2019multi} & 65.92  & 35.14  & 24.17  & 76.85  & 33.93  & 22.78  \\
          & MS + \ours & \textbf{68.15}  & \textbf{36.04}  & \textbf{25.14}  & \textbf{79.74}  & \textbf{34.68}  & \textbf{24.01}  \\\cmidrule{2-8}
          & ProxyAnchor\cite{kim2020proxy} & 67.30  & 37.40  & 26.38  & 84.75  & 37.56  & 27.66  \\
          & ProxyAnchor + \ours & \textbf{67.93}  & \textbf{37.92}  & \textbf{26.88}  & \textbf{87.01}  & \textbf{39.03}  & \textbf{29.39}  \\\bottomrule
    \end{tabular}
    }
  \label{tab:embedding}%
\end{table}%
\paragrapha{4pt}{Effects of embedding size.} Our proposed DIML is also robust across different embedding sizes. Apart from the results in Table~\ref{tab:main} where $D=128$, we perform experiments with $D=64/512$ for Margin~\cite{wu2017sampling}, Multi-Similarity~\cite{wang2019multi} and Proxy Anchor~\cite{kim2020proxy} and the results are summarized in Table~\ref{tab:embedding}. We demonstrate that our method can boost the performance of the three methods consistently no matter how the embedding size $D$ varies.

\subsection{Visualization} To better understand how our method works, we provide some visualizations for CUB200-2011~\cite{cub} in Figure~\ref{fig:vis}, where each pair of images is from the same category. First, we visualize the marginal distributions $\mu^{s}$ and $\mu^t$ (calculated by cross-correlation) through heatmaps and find that they can focus on some discriminative parts in the images (\eg, head, foot, \etc). Second, we show the optimal transport flow $T^*_{i,j}$ and the similarity $S_{i,j}=s(z_i^{\sss}, z_j^{\ttt})$ for some pair of spatial location $(i, j)$. Since the sum of the values in $T^*$ equals 1 and each $T^*_{i,j}$ is relatively small, we use a re-scaled version $\hat{T}^*_{i,j}=G^4T^*_{i,j}$ such that an uniform transport flow yields $\hat{T}^*_{i,j}=1, \forall i,j$. We draw arrows between the location pairs that make a large (or small) contribution to the final structural similarity in red (or blue). The formula along with an arrow takes the form of $\hat{T}^*_{i,j}\times \underline{S_{ij}}$. We observe that our method can match similar parts and assign a higher $T^*_{i,j}$ to the pair while enforcing lower $T^*_{i,j}$ to the parts that are less informative to determine the similarity between the two images. Finally, we demonstrate that by re-weighting the similarity matrix $S$ with the optimal transport matrix $T^*$, our proposed structural similarity (shown in \textbf{\textit{bold}} text) is higher than the standard cosine similarity (shown in \textit{light} text) by a large margin. 

\section{Conclusion}

In this paper, we have presented the deep interpretable metric learning (DIML) method for more transparent embedding learning. We proposed a structural matching strategy that explicitly aligns the spatial embeddings by computing an optimal matching flow between feature maps of the two images.  We evaluated our method on three major benchmarks of deep metric learning including CUB200-2011, Cars196 and Stanford Online Products, and achieved substantial improvements over popular metric learning methods with better interpretability. Our method enables deep models to learn metrics in a more human-friendly way, which can be used to inspect and understand the visual similarity of any two samples or applied to any deep metric learning methods with the proposed multi-scale matching strategy to improve image retrieval performance with controllable cost.

\subsection*{Acknowledgements}

This work was supported in part by the National Natural Science Foundation of China under Grant 61822603, Grant U1813218, and Grant U1713214, in part by a grant from the Beijing Academy of Artificial Intelligence (BAAI), and in part by a grant from the Institute for Guo Qiang, Tsinghua University.

\begin{appendix}
\section{Implementation of DIML}
\subsection{The Sinkhorn Algorithm}
The Sinkhorn algorithm~\cite{cuturi2013sinkhorn} modifies the original optimal transport problem (Eq.4) into the following one:
\begin{equation} \label{eq:sinkhorn}
\begin{split}
& T^* = \argmin_{T \geq 0} \text{tr}(CT^\top) + \lambda \text{tr} \left(T(\log (T) - \mathbf{1} \mathbf{1}^\top)^\top \right) , \\
& \text{subject to} \quad T\mathbf{1} = \mu^\sss, \quad T^\top \mathbf{1} = \mu^\ttt,
\end{split}
\end{equation}
where $\lambda$ is a non-negative regularization parameter. By adding the entropic regularizer, the Equation~\eqref{eq:sinkhorn} becomes a convex problem, which can be solved with Sinkhorn-Knopp algorithm~\cite{sinkhorn1967diagonal}. Starting from an initial matrix $K = \exp({-C}/{\lambda})$, the problem can be solved by iteratively projecting onto the marginal constraints until convergence:
\begin{equation}
    \bm{a} \leftarrow \mu^\sss / K \bm{b}, \quad \bm{b} \leftarrow \mu^\ttt / K^\top \bm{a}.
\end{equation}
After converged, we can obtain the optimal transport plan:
\begin{equation}
    T^* = \text{diag}(\bm{a})K\text{diag}(\bm{b}).
\end{equation}
\subsection{Testing}
In all of our experiments, we use ResNet50~\cite{he2016deep} as our backbone. Therefore, the size of the feature map before the pooling layer is $7\times 7$. To reduce computational costs, we first use ROI Align~\cite{he2017mask} to pool the feature map to $G\times G$ and $G=4$ in most of our experiments unless otherwise noted. According to the multi-scale matching algorithm,  for each image as a query, we first sort the images in the gallery using the standard cosine similarity to obtain the indices of top-$K$ candidates $\mathcal{I}_K$ (we use $K=100$ in most of the experiments). We then calculate the proposed structural similarity of all the images in $\mathcal{I}_K$. To combine both global and structural information, we use the sum of the cosine similarity and the structural similarity for the top-$K$ images to compute their ranks. The regularization parameter $\lambda$ in Equation (6) is set to 0.05.

\subsection{Training}
Incorporating DIML into the training objectives is quite straightforward. Generally, the loss functions in metric learning can be roughly categorized into distance-based methods (\eg, Contrastive~\cite{hadsell2006dimensionality_contrastive}, Triplet~\cite{ge2018hierarchical-triplet}, Margin~\cite{wu2017sampling}) and similarity-based methods (\eg, Multi-Similarity~\cite{wang2019multi}, Arcface~\cite{deng2019arcface}, N-Pair~\cite{sohn2016improved_npair}) For distance-based methods, we replace the original distance function $d$ with the average of $d$ and our structural distance $d_\structural$; For similarity-based methods, we replace the original similarity function $s$ with the average of $s$ and our structural similarity $s_\structural$. In this section, we will use several loss functions as examples to demonstrate how to apply DIML during training.

\paragraph{Margin~\cite{wu2017sampling}} The Margin loss~\cite{wu2017sampling} is defined as
\begin{equation}
    \mathcal{L}_{\rm margin}(k, l) =\left(\sigma+(-1)^{I(y^k\neq y^l)}\left(D_{k,l}-\beta\right)\right)_{+},
\end{equation}
where $\sigma$ and $\beta$ are learnable parameters, and $D_{kl}$ is used to measure the distance between image $k$ and $l$:
\begin{equation}
    D_{k,l} = \frac{1}{2}\left(d_\structural{}(z^k, z^l) + d(\bar{z}^k, \bar{z}^l)\right),
\end{equation}
where $d$ is Euclid distance and $d_\structural$ is derived from $d$ using Equation~(10).

\paragraph{Multi-Similarity~\cite{wang2019multi}} The original Multi-Similarity is defined as:
\begin{equation}
    s^{*}(k, l)=\begin{cases}
s\left(k, l\right), & s\left(k, l\right)>\min _{p \in \mathcal{P}_{k}} s\left(k, p\right)-\epsilon \\
s\left(k, l\right), & s\left(k, l\right)<\max _{n \in \mathcal{N}_{k}} s\left(k, n\right)+\epsilon \\
0, & \text { otherwise }
\end{cases},
\end{equation}
\begin{equation}
\begin{split}
     \mathcal{L}_{\text {MS }}&=\frac{1}{B} \sum_{k \in \mathcal{B}}\left[\frac{1}{\alpha} \log \left[1+\sum_{p \in \mathcal{P}_k} \exp \left(-\alpha\left(s^{*}\left(k, p\right)-\lambda\right)\right)\right]\right.\\
     &\left.+\frac{1}{\beta} \log \left[1+\sum_{n \in \mathcal{N}_{k}} \exp \left(\beta\left(s^{*}\left(k, n\right)-\lambda\right)\right)\right]\right],  
\end{split}
\end{equation}
where $s(k, l)=s(\psi^k, \psi^l)$ is the cosine similarity of the embeddings $\psi^k, \psi^l$ of the two images. To utilize DIML, we can replace $s$ with
\begin{equation}
    s(k, l)\leftarrow \frac{1}{2}\left(s(\bar{z}^k, \bar{z}^l) + s_\structural(z^k, z^l)\right).
\end{equation}
Note that in our notation both $\psi^k$ and $\bar{z}^k$ represent the same embedding in $\mathbb{R}^D$.

\begin{table*}[tbp]
  \centering
  \caption{\textbf{Comparisons of different truncation numbers.} We test for different truncation number $K$ ranging from 0 to 500. Experimental results show that a small $K$ can already bring considerable performance improvement.}
    \begin{tabular}{*{11}{c}}\toprule
 \multirow{2}[4]{*}{\textbf{Baseline}} & \multirow{2}[4]{*}{$K$} & \multicolumn{3}{c}{\textbf{CUB-200}} & \multicolumn{3}{c}{\textbf{Cars196}} & \multicolumn{3}{c}{\textbf{SOP}} \\
\cmidrule(lr){3-5}\cmidrule(lr){6-8}\cmidrule(lr){9-11}            &       & \textbf{P@1} &\textbf{RP} & \textbf{M@R} & \textbf{P@1} & \textbf{RP} & \textbf{M@R} & \textbf{P@1} & \textbf{RP} & \textbf{M@R} \\
\midrule   \multirow{5}[1]{*}{Margin\cite{wu2017sampling}} & 0     & 62.47  & 34.12  & 23.14  & 72.18  & 32.00  & 20.82  & 78.39  & 45.64  & 42.34  \\
          & 10    & \textbf{65.16}  & 34.56  & 23.87  & \textbf{76.65}  & 32.52  & 21.72  & \textbf{79.26} & \textbf{46.44} & \textbf{43.20} \\
          & 50    & 65.16  & 35.43  & \textbf{24.54}  & 76.65  & 33.64  & 22.83  & 79.26  & 46.44  & 43.19  \\
          & 100   & 65.16 & \textbf{35.48} & 24.54 & 76.65 & \textbf{33.93} & \textbf{22.95} & 79.26  & 46.44  & 43.19  \\
          & 500   & 65.16  & 35.48  & 24.54  & 76.65  & 33.93  & 22.95  & 79.26  & 46.44  & 43.19  \\\midrule
    \multirow{5}[0]{*}{Multi-Similarity\cite{wang2019multi}} & 0     & 62.56  & 32.74  & 21.99  & 74.81  & 32.72  & 21.60  & 77.90  & 44.97  & 41.54  \\
          & 10    & \textbf{64.89}  & 33.21  & 22.73  & \textbf{78.50}  & 33.26  & 22.50  & \textbf{78.53} & \textbf{45.60} & \textbf{42.24} \\
          & 50    & 64.89  & 34.04  & 23.37  & 78.50  & 34.46  & 23.72  & 78.53  & 45.60  & 42.23  \\
          & 100   & 64.89 & \textbf{34.12} & \textbf{23.38} & 78.50 & \textbf{34.70} & \textbf{23.81} & 78.53  & 45.60  & 42.23  \\
          & 500   & 64.89  & 34.12  & 23.38  & 78.50  & 34.70  & 23.81  & 78.53  & 45.60  & 42.23  \\\bottomrule
    \end{tabular}%
  \label{tab:trunc_num}%
\end{table*}%

\paragraph{ProxyNCA~\cite{movshovitz2017no_pnca}} It is also worth mentioning there are slight difference when applying DIML to proxy-based methods during training. Taking ProxyNCA~\cite{movshovitz2017no_pnca} as example, the original objective is 
\begin{equation}
    \mathcal{L}_{\text {proxy }}=-\frac{1}{B} \sum_{k \in \mathcal{B}} \log \left(\frac{\exp \left(-d\left(\psi^{k}, \eta^{y^{k}}\right)\right.}{\sum_{c \in \mathcal{C} \backslash\left\{y^k\right\}} \exp \left(-d\left(\psi^k, \eta^c\right)\right.}\right),
\end{equation}

where $d$ is Euclid distance and $\eta^c\in \mathbb{R}^D$ is the proxy for the $c$-th class. To use DIML, we need to use proxies with the size $\mathbb{R}^{H\times W\times D}$, denoted as $\{\rho^c, c\in \mathcal{C}\}$. Then, we can replace the $d(\psi^k, \eta^c)$ with
\begin{equation}
    d(\psi^k, \eta^c)\leftarrow \frac{1}{2}\left(d(\psi^k, \eta^c) + d_\structural(z^k, \rho^c)\right),
\end{equation}
where we also note that $\GAP(\rho^c)=\eta^c$.


\section{Experimental Details}



\subsection{Evaluation Metrics}
We implement the same evaluation metrics as~\cite{musgrave2020metric}, including Precision at 1 (P@1), R-Precision (RP), and Mean Average Precision at R (MAP@R).

\noindent \textbf{P@1} is also known as Recall@1 in metric learning. Given a sample $x^q$ and feature encoder $\phi(\cdot)$, the set of $k$ nearest neighbors of $x^q$ is calculated as the precision of $k$ nearest neighbors:
\begin{equation}
    \mathcal{N}_q^k = \argmin_{\mathcal{N}\subset\mathcal{X_{\text{test}}},|\mathcal{N}|=k} \sum_{x^f\in\mathcal{N}} d_e(\phi(x^q),\phi(x^f))
    \label{k-neighbor}
\end{equation}
where $d_e(\cdot, \cdot)$ is the euclidean distance. Then P@$k$ can be measured as
\begin{equation}
    \text{P@}k =  \frac{1}{|\mathcal{X_{\text{test}}}|}\sum_{x_q\in \mathcal{X}_{\text{test}}}\frac{1}{k}\sum_{x^i\in \mathcal{N}_q^k}\begin{cases} 
    1,\quad y^i = y^q,\\
    0,\quad \text{otherwise}
    \end{cases},
    \label{equ:precision}
\end{equation}
where $y^i$ is the class label of sample $x^i$. We only report P@1 in our experiments, i.e. $k=1$.

\noindent \textbf{R-precision} is defined in~\cite{musgrave2020metric}. Specifically, for each sample $x^q$, let $R$ be the number of images that are the same class with $x^q$ and R-precision is simply defined as P@$R$ (see Equation~\ref{equ:precision}).
 However, R-precision does not consider the ranking of correct retrievals, so it is not informative enough. To tackle this problem, ~\cite{musgrave2020metric} introduced Mean Average Precision at R.

\noindent \textbf{MAP@R} is similar to mean average precision, but limit the number of nearest neighbors to R. So it replaces \textit{precision} in MAP calculation with \textit{R-precision}:
\begin{equation}
    \text{MAP}@R = \frac{1}{R}\sum_{i=1}^{R} P(i),
\end{equation}
where
\begin{equation}
    P(i)=\left\{\begin{array}{ll}
\text {P@} i, & \text { if the } i\text{-th} \text{ retrieval is correct; } \\
0, & \text { otherwise. }
\end{array}\right.
\end{equation}
\begin{table*}[htbp]
  \centering
  \caption{\textbf{Effects of the size of feature map.} Generally, the performance of our \ours is better with higher $G$. \ours~with $G=4$ yields good results within relatively low computational costs.}\label{tab:fm_size}
  \adjustbox{max width=\textwidth}{
    \begin{tabular}{*{11}{c}}\toprule
    \multirow{2}[0]{*}{\textbf{Baseline}} & \multirow{2}[0]{*}{$G$} & \multicolumn{3}{c}{\textbf{CUB-200}} & \multicolumn{3}{c}{\textbf{Cars196}} & \multicolumn{3}{c}{\textbf{SOP}} \\\cmidrule(lr){3-5}\cmidrule(lr){6-8}\cmidrule(lr){9-11}
          &       & \multicolumn{1}{c}{\textbf{P@1}} & \textbf{RP} & \textbf{M@R} & \textbf{P@1} & \textbf{RP} & \textbf{M@R} & \textbf{P@1} & \textbf{RP} & \textbf{M@R} \\\midrule
    \multirow{4}[0]{*}{Margin\cite{wu2017sampling}} & 1     & 62.47  & 34.12  & 23.14  & 72.18  & 32.00  & 20.82  & 78.39  & 45.64  & 42.34  \\
          & 2     & 64.15  & 34.79  & 23.83  & 75.04  & 32.59  & 21.85  & 79.06  & 46.29  & 43.03  \\
          & 4     & 65.16  & 35.48  & 24.54  & 76.65  & \textbf{33.93}  & \textbf{22.95}  & 79.26  & 46.44  & 43.19  \\
          & 7     & \textbf{65.58}  & \textbf{35.58}  & \textbf{24.79}  & \textbf{76.96}  & 32.93  & 22.66  & \textbf{79.59}  & \textbf{46.83}  & \textbf{43.62}  \\\midrule
    \multirow{4}[0]{*}{Multi-Similarity~\cite{wang2019multi}} & 1     & 62.56  & 32.74  & 21.99  & 74.81  & 32.72  & 21.60  & 77.90  & 44.97  & 41.54  \\
          & 2     & 63.77  & 33.33  & 22.60  & 77.45  & 33.25  & 22.60  & 78.39  & 45.56  & 42.15  \\
          & 4     & 64.89  & 34.12  & 23.38  & 78.50  & \textbf{34.70}  & \textbf{23.81}  & 78.53  & 45.60  & 42.23  \\
          & 7     & \textbf{65.45}  & \textbf{34.15}  & \textbf{23.55}  & \textbf{78.93}  & 33.64  & 23.50  & \textbf{78.76}  & \textbf{45.90}  & \textbf{42.57}  \\\bottomrule
    \end{tabular}%
    }
  \label{tab:grid_size}%
\end{table*}%

MAP@R is more informative than P@1 and it can be computed directly from the embedding space without clustering as post-processing. 

\subsection{Experimental Setups}

For most of the baseline methods, we follow the implementation and the hyper-parameters in~\cite{szegedy2016rethinking}. For Proxy Anchor~\cite{kim2020proxy}, we use their original implementation but set the hyper-parameters as~\cite{szegedy2016rethinking} (batch size 112, embedding size 128, \etc). Besides various loss functions, we also experiment with different sampling methods. In Table 1 of the original paper, we use suffixes to represent the sampling methods (-R: Random; -D: Distance~\cite{wu2017sampling}; -S Semihard~\cite{schroff2015facenet}; -H: Softhard~\cite{roth2020revisiting}). 

\section{Detailed Results}

In the original paper, we have demonstrated the effects of truncation number $K$ and feature map size $G$ using charts. In this section, we provide the original numerical results that were used to plot those charts in Table~\ref{tab:trunc_num} and Table~\ref{tab:fm_size}.

\end{appendix}
{\small
\bibliographystyle{ieee_fullname}
\bibliography{reference,dml,xai}
}
\end{document}